%% file: arxiv.tex
\pdfoutput=1

\documentclass{article}
\usepackage[preprint]{neurips_2023}

\usepackage[utf8]{inputenc}
\usepackage[T1]{fontenc}
\usepackage{url}
\usepackage{amsfonts}
\usepackage{nicefrac}
\usepackage{microtype}
\usepackage{booktabs}
\input{math_commands}

\usepackage{paralist}
\usepackage{wrapfig}

\usepackage{xspace}

\usepackage{tabulary}
\usepackage{color, colortbl}
\definecolor{ForestGreen}{RGB}{34, 139, 34}
\definecolor{bleudefrance}{rgb}{0.19, 0.55, 0.91}
\usepackage[colorlinks=true,linkcolor=orange,citecolor=bleudefrance]{hyperref}
\usepackage[normalem]{ulem}
\usepackage{multirow}
\usepackage{float}
\usepackage{paralist}
\usepackage{comment}
\usepackage{subcaption}

\usepackage[textwidth=0.8in,textsize=tiny]{todonotes}
\usepackage{soul}

\usepackage{makecell}

\usepackage{array}
\newcolumntype{H}{>{\setbox0=\hbox\bgroup}c<{\egroup}@{}}

\usepackage{graphicx}
\usepackage{amsmath}
\usepackage{amssymb}
\usepackage{amsthm}
\usepackage{mathtools}
\usepackage{geometry}
\usepackage{fancyhdr}
\usepackage{setspace}
\usepackage{natbib}

\definecolor{mycolor}{RGB}{0,0,255}

\usepackage[capitalize]{cleveref}
\crefformat{equation}{Eq.~(#2#1#3)}
\crefname{section}{§}{§§}
\Crefname{section}{§}{§§}

\usepackage{tcolorbox}

\theoremstyle{plain}

\title{What Makes the Preferred Thinking Direction for LLMs in Multiple-choice Questions?}

\author{
 \textbf{Yizhe Zhang\textsuperscript{1}}\thanks{Equal contribution.},\quad 
 \textbf{Richard He Bai\textsuperscript{1}}\footnotemark[1],\quad 
 \textbf{Zijin Gu\textsuperscript{1}}\thanks{Core contribution.},\quad
 \textbf{Ruixiang Zhang\textsuperscript{1}},\quad 
\\
 \textbf{Jiatao Gu\textsuperscript{1}},\quad
 \textbf{Emmanuel Abbe\textsuperscript{1}},\quad 
 \textbf{Samy Bengio\textsuperscript{1}},\quad 
 \textbf{Navdeep Jaitly\textsuperscript{1}}
\\
\\
 \textsuperscript{1}Apple
\\
 \small{
{\{yizhe\_zhang, richardbai, njaitly\}@apple.com}
 }
}

\begin{document}

\maketitle
\begin{abstract}
Language models usually use left-to-right (L2R) autoregressive factorization.
However, L2R factorization may not always be the best inductive bias.
Therefore, we investigate whether alternative factorizations of the text distribution could be beneficial in some tasks.
We investigate right-to-left (R2L) training as a compelling alternative, focusing on multiple-choice questions (MCQs) as a test bed for knowledge extraction and reasoning. Through extensive experiments across various model sizes (2B-8B parameters) and training datasets, we find that R2L models can significantly outperform L2R models on several MCQ benchmarks, including logical reasoning, commonsense understanding, and truthfulness assessment tasks. Our analysis reveals that this performance difference may be fundamentally linked to multiple factors including calibration, computability, and directional conditional entropy.
We analyze the impact of these factors through controlled simulation studies using arithmetic tasks, where the impacting factors can be better disentangled. 
Our work demonstrates that exploring alternative factorizations of the text distribution can lead to improvements in LLM capabilities and provides theoretical insights into optimal factorization towards approximating human language distribution, and when each reasoning order might be more advantageous. Our code and checkpoints are released at \url{https://github.com/apple/ml-reversal-blessing}
\end{abstract}

\begin{figure*}[ht!]
    \centering
    \includegraphics[width=0.99\linewidth]{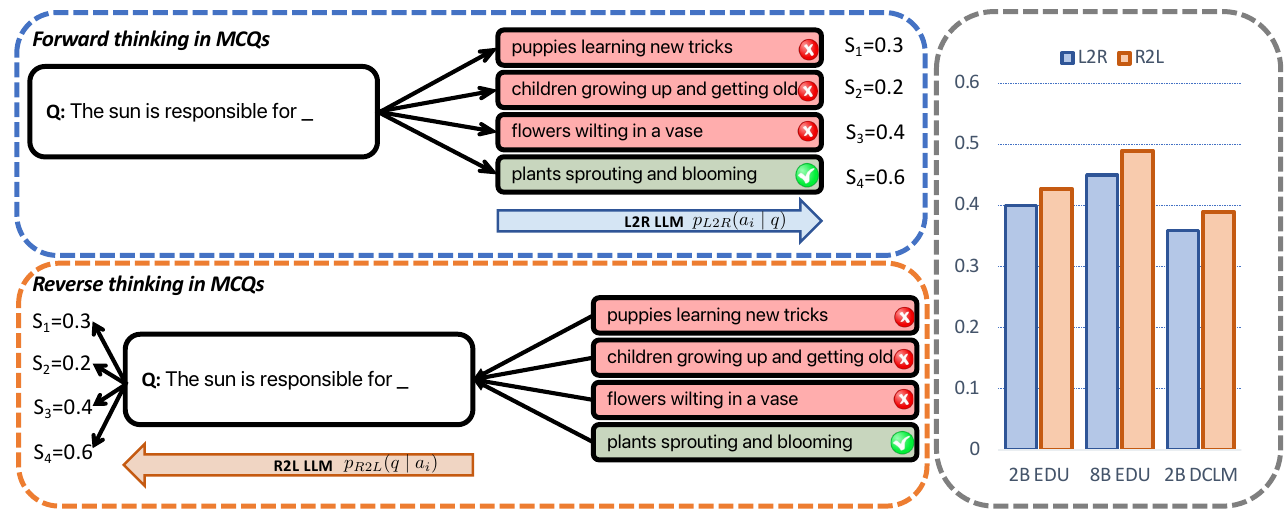}
    \caption{Reverse Thinking in MCQs.
\textbf{Top left}: Standard forward thinking evaluates each answer choice based on the question and selects the one with the highest relevance score in a L2R LLM.
\textbf{Bottom left}: Reverse thinking evaluates the question based on each answer choice and selects the answer that maximizes the relevance score in a R2L LLM.
\textbf{Right}: Reverse thinking consistently outperforms forward thinking in certain MCQ tasks (Openbook QA in this figure), independent of training data and model size.
}
    \label{fig:r2l}
\end{figure*}

\section{Introduction}
Large Language Model (LLM) pretraining commonly employs left-to-right (L2R) next-token prediction, an approach that enables efficient parallelization and caching. This method models the text distribution $p(x)$ as a factorized autoregressive chain as $p(x_t|x_{<t})$. L2R naturally aligns with human cognitive processes of text generation and reasoning, making it well-suited for inference tasks.

However, while perfect modeling of each $p(x_t|x_{<t})$ would theoretically enable exact recovery of the data distribution $p(x)$, neural networks inevitably introduce approximation errors for each $p(x_t|x_{<t})$.
These errors compound over timestep $t$ during inference, potentially resulting in hallucinations and repetitions in generation \citep{bengio2015scheduled, zhang2023planner}.
Further, L2R factorization can result in inductive biases that lead to unwanted behaviors.
For example, \citet{allen2023physics_3_1} show that inverse search is challenging for L2R LLMs, and \citet{berglund2023reversal} demonstrate the "reversal curse" where models trained on forward text data struggle with inverse relationships.

We investigate whether L2R is optimal, and if alternative factorizations might capture unique aspects of the data distribution that complement L2R. Can specific factorizations achieve lower approximation errors compared to L2R, or reduce L2R's inherent bias in particular task domains? 

Autoregressive modeling in right-to-left (R2L) fashion factorizes $p(x)$ as $p(x_t|x_{>t})$, which presents a particularly promising alternative that has been examined in previous work \citep{papadopoulos2024arrows, berglund2023reversal, zhang2024reverse}. 
This setup views the task as predicting the previous token, and it can achieve prediction losses comparable to the L2R next token prediction objective, due to its symmetry to L2R.
While R2L may seem counterintuitive given human language processing patterns, it may enable more efficient knowledge extraction in certain scenarios by aligning with the natural direction of information flow in those cases, and it could provide complementary inductive biases that help with specific reasoning tasks.

We investigate three questions: (1) How to evaluate R2L models on knowledge extraction and basic reasoning tasks? (2) Can R2L factorization match or surpass L2R's capabilities in knowledge extraction and reasoning for downstream tasks? (3) What are the underlying factors determining the preference of L2R or R2L factorizations?

To address these questions, we conducted controlled experiments comparing L2R and R2L models trained with identical data and computational resources.
We evaluated both factorization approaches using standard LLM benchmarks with Multiple-Choice Questions (MCQs). For simplicity, we limit our comparison to MCQs, and leave the evaluations for generative tasks as future work.
For R2L models, we applied Bayesian inference to implement "reverse thinking," evaluating choices based on their likelihood of generating the prompt (Figure~\ref{fig:r2l}).

Our results reveal a surprising and previously unobserved empirical finding: R2L models can significantly outperform L2R models on several standard MCQ benchmarks, including tasks requiring logical reasoning, commonsense understanding, and truthfulness assessment. This finding—where backward thinking can sometimes yield superior performance—is a phenomenon not previously observed or widely recognized in the current literature, especially at the scale and across the diverse benchmarks we evaluated. This observation challenges the prevailing fundamental assumption that L2R autoregressive factorization is the default and optimal inductive bias for LLMs.

Beyond this empirical discovery, our work introduces new perspectives on analyzing the performance differences between L2R and R2L models by proposing three potential underlying factors: \textit{calibration}, \textit{computability}, and, critically, \textit{conditional entropy}. The role of conditional entropy in explaining the preferred reasoning direction, and generally understanding reasoning machinery, is a novel theoretical insight introduced in this paper. We empirically verify this hypothesis, showing that lower conditional entropy generally correlates with higher accuracy in the reasoning direction for many tasks. Our primary contribution lies not in demonstrating that one factorization is universally superior, but in developing a principled framework for understanding when and why different factorization directions may be preferred.

Nevertheless, these factors are intricately interwoven in actual MCQs, complicating the analysis. To disentangle these factors and ablate on their impact to the performance of L2R or R2L factorization, we design a controlled simulation study using arithmetic tasks, revealing how various factors influence the effectiveness of certain factorization.
Our code and model checkpoints have been made publicly available for reproduction and to facilitate future research.
\footnote{\scriptsize \url{https://github.com/apple/ml-reversal-blessing}.}

\section{Thinking Backward in MCQs}
\subsection{Solving MCQs}
\label{sec:rev}
\paragraph{Solving MCQs with forward thinking}
As shown in Figure~\ref{fig:r2l}, in MCQs, LLMs process a question $q$ alongside a set of answer choices $ A = \{a_1, a_2, \ldots, a_n\} $. Each (question, answer) pair $(q, a_i)$ is encoded to compute a relevance score $s_i$. The model then selects the answer $a_k$ corresponding to the highest score: $k = \arg\max_i s_i$.

To compute $s_i$, the model evaluates the log-probability of generating the answer $a_i$ given the question $q$. This log-probability is often normalized to account for variations in answer length, preventing a bias toward shorter or longer responses. Various normalization techniques \citep{holtzman-etal-2021-surface} can be applied, however, we resort to the most common approach which divides the total log-probability by the length of the answer $N_i=\text{len}(a_i)$ in tokens or bytes, resulting in a normalized relevance score:
$s_i = \frac{\log p(a_i \mid q)}{N_i}$. The log-probability is factorized as 
\begin{align}
    \log p(a_i \mid q) = \sum_{l=1}^{N_i} \log p_{L2R}(a_i^{l} \mid q, a_i^{<l}),
    \label{eq:forward}
\end{align}
where $a_i^{l}$ represents the $l$-th token in $a_i$.

\paragraph{Solving MCQs with reverse thinking}
With an R2L model, $s_i$ can be computed using Bayes' rule:
\begin{align}
    s_i &= \log p(a_i \mid q) / M_i = \frac{1}{M_i}(\log p_{R2L}(q \mid a_i) + \log p_{R2L}(a_i) - C), \nonumber
\end{align}
where $M_i=\text{len}(q,a_i)$,  $C=\log p_{R2L}(q)$ is a constant. $\log p_{R2L}(q \mid a_i)$ and $\log p_{R2L}(a_i)$ can be autoregressively factorized in R2L manner similar to the forward thinking process in Eq.~\eqref{eq:forward}. We consider 3 paradigms of the $s_i$ for reverse thinking: (1) normalized $s_i$ with $M_i =\text{len}(q, a_i)$ resembling the forward thinking; (2) unnormalized $s_i$ with $M_i = 1$; (3) unnormalized $s_i$ without prior, i.e. $s_i=\log p_{R2L}(q \mid a_i)$.

\subsection{Model evaluation}
We conducted our evaluation on standard LLM evaluation tasks with MCQs that cover different domains including commonsense reasoning, logical reasoning, truthfulness evaluation and more. 

Our evaluation tasks include HellaSwag~\citep{zellers-etal-2019-hellaswag}, ARC~\citep{clark2018think}, MMLU~\citep{hendrycks2021measuring}, Openbook QA~\citep{mihaylov2018openbookqa}, MathQA~\citep{amini-etal-2019-mathqa}, LogiQA~\citep{liu2020logiqa}, PIQA~\citep{bisk2019piqa}, Social IQA~\citep{sap-etal-2019-social}, Commonsense QA \citep{talmor2018commonsenseqa}, Truthful QA~\citep{lin2021truthfulqa}, and WinoGrande~\citep{sakaguchi2021winogrande}. 
For ARC (easy, hard) and MMLU, we combine all the subtasks to report the overall score. 
We use Eleuther-AI LM-eval harness \citep{eval-harness} for all the evaluations. 
For MMLU, LogiQA, and Commonsense QA, we modify the task templates to present full answer choices rather than just choice labels.

\begin{table*}[!htp]\centering
\caption{Comparing L2R and R2L on MCQs. All the models are trained on 350B non-repeating tokens. The HF-2B baseline is from \citet{penedo2024the}. We directly used their reported numbers. EDU-2B, EDU-8B and HF-2B models are trained with the same FineWeb-EDU 350B dataset. \textcolor{ForestGreen}{Green} indicates R2L wins, \textcolor{red}{red} indicates R2L loses.
}\label{tab:main_results}
\rowcolors{2}{gray!15}{white}
\scriptsize
\begin{tabular}{lcccccccccccc}
\toprule
&\multicolumn{3}{c}{\textbf{DCLM-2B}} &\multicolumn{3}{c}{\textbf{EDU-2B}} &\multicolumn{3}{c}{\textbf{EDU-8B}} & \textbf{HF-2B} \\\cmidrule{2-11}
&L2R &R2L &\% Change &L2R &R2L &\% Change &L2R &R2L &\% Change &L2R \\
\midrule
Training loss & \textbf{2.668} & 2.724 & \textcolor{red}{+2.10} &\textbf{2.345} &2.396 & \textcolor{red}{+2.17} & \textbf{2.087}& 2.138 & \textcolor{red}{+2.44} & - \\
\midrule
\textbf{LogiQA} &30.57 &\textbf{31.64} & \textcolor{ForestGreen}{+3.52} &27.96 &\textbf{31.49} & \textcolor{ForestGreen}{+12.64} &29.95 &\textbf{31.03} & \textcolor{ForestGreen}{+3.61} & - \\
\textbf{OpenbookQA} &36.00 &\textbf{38.40} & \textcolor{ForestGreen}{+6.67} &42.40 &\textbf{44.40} & \textcolor{ForestGreen}{+4.72} &45.00 &\textbf{48.40} & \textcolor{ForestGreen}{+7.56} & 41.04 \\
\textbf{TruthfulQA} &19.82 &\textbf{29.99} & \textcolor{ForestGreen}{+51.23} &24.36 &\textbf{28.76} & \textcolor{ForestGreen}{+18.09} &24.97 &\textbf{31.70} & \textcolor{ForestGreen}{+26.95} & - \\
\textbf{CommonsenseQA} &42.83 &\textbf{45.29} & \textcolor{ForestGreen}{+5.74} &42.92 &\textbf{45.13} & \textcolor{ForestGreen}{+5.15} &39.15 &\textbf{44.96} & \textcolor{ForestGreen}{+14.84} & 36.60 \\
Social IQA &\textbf{41.56} &40.94 & \textcolor{red}{-1.48} &\textbf{42.78} &42.22 & \textcolor{red}{-1.32} &\textbf{44.58} &43.50 & \textcolor{red}{-2.42} & 40.52 \\
ARC &\textbf{54.11} &43.88 & \textcolor{red}{-18.91} &\textbf{60.65} &52.31 & \textcolor{red}{-13.75} &\textbf{68.29} &56.22 & \textcolor{red}{-17.67} & 57.47 \\
HellaSwag &\textbf{60.87} &45.89 & \textcolor{red}{-24.62} &\textbf{60.57} &44.34 & \textcolor{red}{-26.79} &\textbf{71.60} &49.22 & \textcolor{red}{-31.26} & 59.34 \\
MathQA &\textbf{26.50} &22.21 & \textcolor{red}{-16.18} &\textbf{26.80} &24.86 & \textcolor{red}{-7.25} &\textbf{28.77} &25.33 & \textcolor{red}{-11.96} & - \\
MMLU &\textbf{31.66} &31.31 & \textcolor{red}{-1.10} &\textbf{34.57} &34.35 & \textcolor{red}{-0.62} &\textbf{38.90} &37.11 & \textcolor{red}{-4.60} & 37.35 \\
PIQA &\textbf{74.43} &58.05 & \textcolor{red}{-22.00} &\textbf{74.48} &57.13 & \textcolor{red}{-23.30} &\textbf{77.80} &59.14 & \textcolor{red}{-23.98} & 76.70 \\
Winogrande &\textbf{61.01} &53.51 & \textcolor{red}{-12.29} &\textbf{60.93} &54.85 & \textcolor{red}{-9.97} &\textbf{65.75} &54.70 & \textcolor{red}{-16.81} & 57.54 \\
\bottomrule
\end{tabular}
\end{table*}

\subsection{Model Pretraining}
To pretrain the model, we first tokenize each complete dataset. 
The R2L model is then trained by reversing all tokens within each training data instance. 
For a fair comparison between the R2L and L2R models, both models are pretrained from scratch using the same Fineweb-EDU subset dataset comprising 350B tokens \citep{penedo2024the}. 
Each model consists of 2B parameters (\textbf{EDU-2B}), which is the default setting in our experiments. 
We also train 1.5B, 4B, 8B L2R and R2L models with the same 350B Fineweb-EDU dataset, and 2B L2R and R2L models trained with a random subset of the DCLM dataset \citep{li2024datacomplm} containing 350B tokens (\textbf{DCLM-2B}).
Both the L2R and R2L models are trained for a single epoch, ensuring each training instance is seen only once, thus the training loss should align with the validation loss. More details for model architecture and training are provided in Appendix~\ref{app:arch}.

\subsection{Results}
We present our results in Table~\ref{tab:main_results}. To verify our pretraining pipeline, we first compare the performance of our pretrained model with the 2B model trained by Huggingface \citep{penedo2024the} (\textbf{HF-2B}) 
\footnote{\url{https://huggingface.co/spaces/HuggingFaceFW/blogpost-fineweb-v1}}. 
Under similar model size and the same dataset, our 2B model (\textbf{EDU-2B}) achieves performance comparable to or exceeding the L2R results reported by Huggingface \textbf{HF-2B}.
Full results for all models ranging from 1.5B to 8B parameters are provided in our appendix~\ref{app:full_results}. 

We compared L2R and R2L model performance across all evaluated tasks, employing bootstrap sampling (5 replicates, each with 80\% resampling with replacement) for statistical robustness. As shown in Table~\ref{tab:main_results}, R2L models with reverse thinking exhibited significantly better reasoning performance on 4 out of 11 tasks: LogiQA, OpenBookQA, TruthfulQA, and CommonsenseQA. Statistical significance results are presented in Table~\ref{tab:ttest}.

These results remained consistent across different model sizes (1.5B to 8B), datasets (DCLM, FineWeb EDU), and random seeds, indicating the findings are not due to random fluctuation. The relative performance gain or loss when switching to R2L remained generally stable as model size increased (see appendix~\ref{app:full_results}).

For TruthfulQA specifically, we observed the most significant performance gain with R2L the improvement was substantial (51.23\% on DCLM-2B). We hypothesize that the "reverse thinking" may inherently align better with truthfulness assessment, as it evaluates the question based on each answer choice rather than generating answers from the question. This framing might help the R2L model better discern subtle inaccuracies that an L2R model might overlook due to "surface form competition". Additionally, R2L models demonstrate significantly lower conditional entropy on TruthfulQA compared to L2R models, which aligns with our hypothesis that lower conditional entropy is associated with higher task accuracy.

For reverse thinking with R2L, we use the paradigm 3 (\textit{i.e.}, unnormalized $s_i$ without prior) for downstream tasks evaluation. 
We compare the three paradigms for reverse thinking in Appendix~\ref{app:rev_comparison}, Table~\ref{tab:3variants}.
Ideally, $s_i$ should incorporate priors, as in paradigm 1 or 2. However, in practice, using $s_i$ without prior (paradigm 3) consistently yields the best performance except for Social IQA and PIQA. We hypothesize this may be due to intrinsic difficulty of estimating the prior probabilities $p(a)$ using LLMs, due to the "surface competition" calibration issues \citep{holtzman-etal-2021-surface}. We provide detailed explanation of our hypothesize using an illustrative example in Appendix~\ref{app:surface}. 

Intuitively, paradigm 3 which uses $p(q \mid a)$ is sensible.
In MCQs, answer choices are typically well-formed and reasonable text, meaning their prior probabilities $p(a)$ are unlikely to vary significantly among choices, assigning a uniform prior is probably a reasonable approach. 
Consequently, $p(a \mid q)$ and $p(q \mid a)$ tend to be highly correlated. Consider a real-world example from Openbook QA: for the question $(q)$ "A magnet will stick to", candidate answers $(a_i)$ include "a belt buckle", "a wooden table", "a plastic cup", and "a paper plate". A model can deduce that "a belt buckle" is far more likely to be associated with the question "A magnet will stick to" compared to the other options, demonstrating how $p(q \mid a)$ can effectively capture the relevance between question and answer.

We also monitor the training loss for pretraining the models on both directions.
We observed findings similar to~\citet{papadopoulos2024arrows} in that L2R yields a lower loss compared to R2L, even though both models model the same target data distribution.
In \citet{papadopoulos2024arrows}, the largest model that was trained had $405$M parameters while our models were trained at the popular small LLM size range of 2B-8B parameters.
At this size, we observe a similar percentage difference as reported by previous work, of about 2\%-2.5\% increase in loss when using R2L, indicating learning the R2L factorization is more challenging. This makes it particularly interesting that on a bunch of MCQ tasks we see the R2L is performing better, as elaborated above.

\begin{figure*}[htp!]
    \centering
    \includegraphics[width=1.0\linewidth]{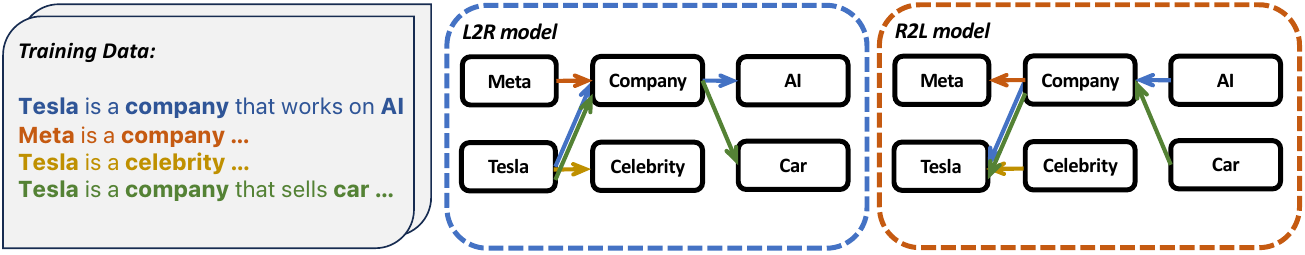}
    \caption{L2R and R2L LLMs pretrained on the same data will generate opposite search graphs based on the order in which they process the information entities.}
    \label{fig:graph}
\end{figure*}

\section{What Makes the Preferred Order of Thinking?}
\label{sec:why}
We then seek to gain a deeper understanding of why there is a preferred orientation for the MCQs. We explore three main hypotheses (3\textbf{C}): \textit{\textbf{C}alibration}, \textit{\textbf{C}omputability}, and \textit{\textbf{C}onditional entropy}. 
Admittedly, there may be other factors that we have overlooked that contributes to this preference. 

\subsection{Calibration} The first potential explanation concerns the scoring mechanism in forward thinking, where $s_i=\log p_{L2R}(a_i \mid q)$. Eq.~\eqref{eq:forward} might not lead to an optimal estimation of $p(a|q)$ as it suffers from several calibration issues. Among the choices, some may contain more words that are highly predictable (e.g., "Hong Kong" or stop-words like "a"), potentially leading to spuriously inflated relevance scores. Additionally, \citet{holtzman-etal-2021-surface} shows that simple probability normalization in MCQs is challenging because different surface forms of semantically equivalent answers compete for probability mass, potentially \textit{diluting} scores for correct answers due to this "surface form competition".

In contrast, reverse thinking with paradigm 3, where $s_i=\log p_{R2L}(q \mid a_i)$, mitigates this issue since the target question $q$ remains constant across all choices.
We provide rationale analysis on how R2L paradigm 3 alleviates "surface competition" in Appendix~\ref{app:surface}.
In a nutshell, forward thinking suffers from surface form competition, where semantically similar words (e.g., "dog" and "puppy") split probability mass, reducing the likelihood of selecting the correct answer. Reverse thinking mitigates this by enforcing a uniform prior, eliminating competition in the prior distribution and allowing a fairer comparison between answer choices.
This suggests that reverse thinking inherently "auto-normalizes" different choices, resulting in more robust evaluation. However, this sole theory fails to explain why reverse thinking does not consistently outperform forward thinking across all tasks, instead showing superior performance only in specific MCQ scenarios.

\subsection{Computability} A second potential theoretical explanation, which echoes with \citet{papadopoulos2024arrows}, suggests that computational complexity may underlie these directional preferences. Drawing an analogy to number theory, where multiplying prime numbers is computationally straightforward, while the reverse operation of prime factorization is NP-hard. 

It is tempting to consider this computational complexity asymmetry as the main underlying cause for why L2R or R2L is preferred for specific tasks. However,
recent research \citep{mirzadeh2024gsm,kambhampati2024can,valmeekam2024llms} finds that LLMs may not actually perform genuine reasoning or computing,
as evidenced by their poor generalization when tasks undergo minor modifications. 
This implies that LLMs mainly emulate \textit{reasoning patterns} from their training data instead of carrying out actual logical computation, weakening the hypothesis that directional preferences stem from varying computability in different directions. Furthermore, most MCQs primarily involve knowledge retrieval and basic reasoning, which might not reach the complexity threshold where computational hardness would become a significant factor. Therefore, acknowledging that computability may be a factor, we keep exploring alternative hypotheses.

\begin{figure*}[ht!]
    \centering
    \includegraphics[width=0.8\linewidth]{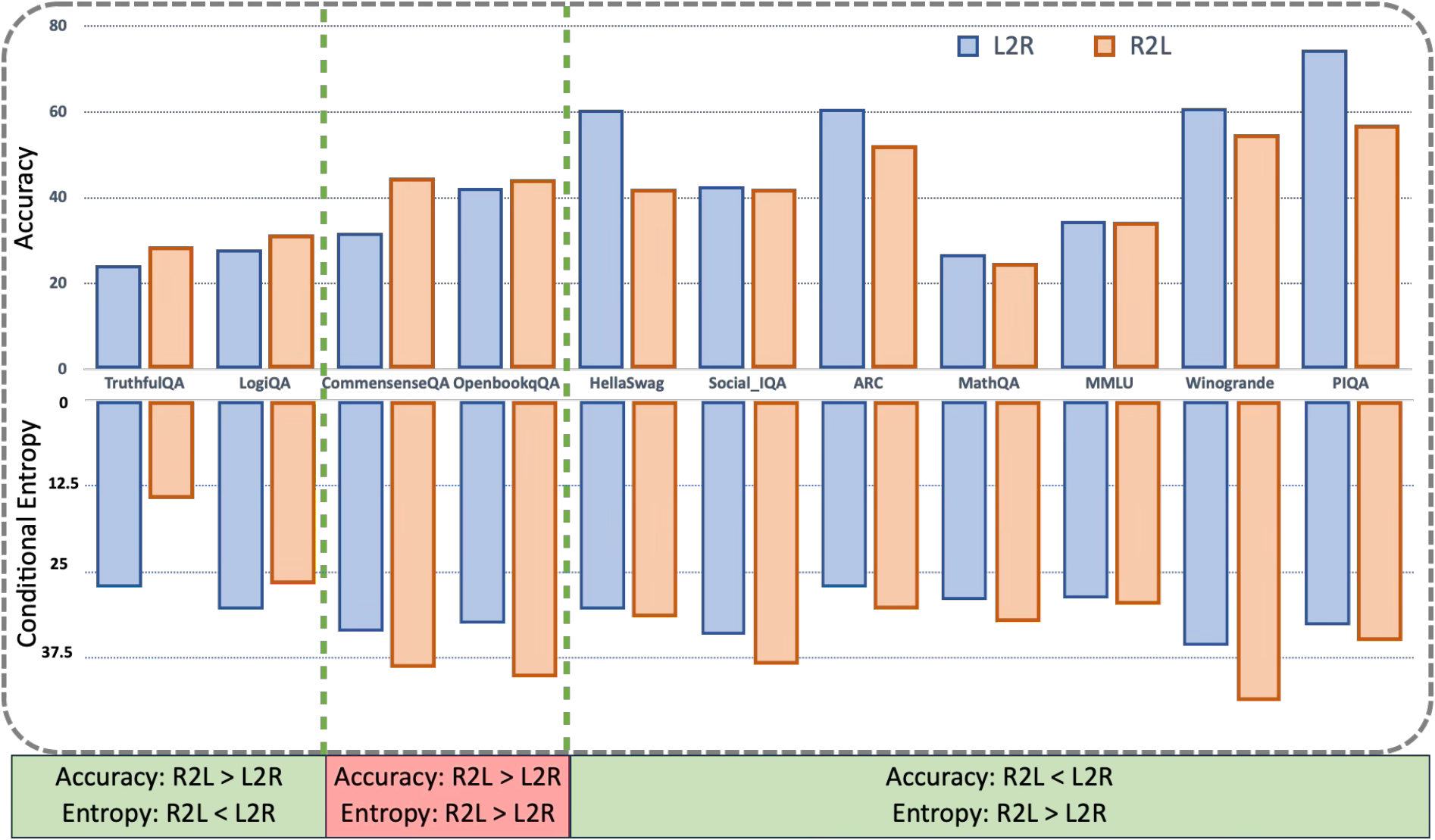}
    \caption{
    Lower conditional entropy may associate with higher accuracy in the reasoning direction.
    }
    \label{fig:ppl_comparison}
\end{figure*}

\subsection{Conditional Entropy}
Our final hypothesis posits that the optimal direction of thinking is closely related to the \textit{conditional entropy} of the downstream task. 
Recent work has shown that learning knowledge extraction and simple multihop reasoning is more challenging for problems with higher degree of \textbf{branching factors} or "\textbf{globality degree}" compared to those with lower branching factors and more deterministic relationships~\citep{abbe2024far}.
It is conceivable that directionality of data can impact the branching degree and lead to different learning efficiencies in different directions (for example multiplication in left-to-right direction is factorization in the opposite direction, each with different branching factors).

Previous work \citep{berglund2023reversal,allen2023physics_3_2} has also demonstrated that LLMs suffer from the "reversal curse", indicating that inverse R2L search in LLMs is inherently challenging for L2R models - due the disconnect between training and inference directions.
Consider an LLM trained on sequences of knowledge/information name entities $(e_1,e_2,\cdots,e_n)$. LLM may effectively construct a \textbf{directed} search graph that maps the key $(e_1,\cdots,e_{i-1})$ to the value $e_i$ for any $i$.  Following this logic, the training data essentially forms a Bayesian network that can be represented as a \textit{directed acyclic graph} (DAG) of entities. Similarly, training an R2L model yields an analogous DAG but with reversed edge directions (see Figure~\ref{fig:graph} for an illustration).
The search efficiency between these two graphs may vary given different queries.

We hypothesize here that for two different factorizations of the data, \textbf{the direction yielding lower conditional entropy will perform better in MCQs}, as it reflects better efficiency in knowledge extraction and multi-hop search. 
We note however, that this is only true when models under both factorization directions have sufficiently low error, which seems to be true for our models here.

More formally, for a downstream MCQ task $T$ with question and answer choices following task-specific data distribution $P_T(q, a)$, we compare the \textit{conditional entropy} in both directions under pretrained L2R and R2L models (Eq.~\eqref{eq:l2r_ent} for L2R and Eq.~\eqref{eq:r2l_ent} for R2L):
\begin{align}
&-\mathbb{E}_{q'\sim P_T(q)} \sum_a p_{L2R}(a|q') \log p_{L2R}(a|q').
\label{eq:l2r_ent}\\
&-\mathbb{E}_{a'\sim P_T(a)} \sum_q p_{R2L}(q|a') \log p_{R2L}(q|a').
\label{eq:r2l_ent}
\end{align}

We assume that the conditional entropy is a proxy for the quality of the learned model, and the direction with lower conditional entropy should perform better.
However, computing these summations in \eqref{eq:l2r_ent} and \eqref{eq:r2l_ent} is intractable due to the exponentially large candidate space. Therefore, we employ Monte Carlo estimation of \eqref{eq:l2r_ent} and \eqref{eq:r2l_ent}
as proxy measures, specifically computing \begin{align}
&-\mathbb{E}_{q' \sim P_t(q), a' \sim p_{L2R}(a|q')} \log p_{L2R}(a' | q'), \\
&-\mathbb{E}_{a' \sim P_t(a), q' \sim p_{R2L}(q|a')} \log p_{R2L}(q' | a'). 
\end{align}

Because of the extensive amount of evaluation datasets, due to limited computation budget, we only conducted a single sample rollout for $a' \sim p_{L2R}(a|q')$ and $q' \sim p_{R2L}(q|a')$. We recognize that this may not be a precise representation of the true conditional entropy, given that the candidate space grows exponentially with the maximum sequence length.

\paragraph{Empirical Verification}
To verify this hypothesis, we estimate the conditional entropy for all the evaluation tasks. We provide more experimental details in Appendix~\ref{app:ce}.
Figure \ref{fig:ppl_comparison} presents our empirical results, which support this hypothesis that
lower conditional entropy is typically linked to greater task accuracy, except for CommonsenseQA and OpenbookQA which are outliers likely because of other confounding factors including the computability. 
In Figure \ref{fig:ppl_comparison}, we observed that the conditional entropy of R2L is generally greater than L2R. This trend could be related to the findings presented in Table~\ref{tab:main_results}, indicating that R2L tends to have higher training loss too.
Complementing the rationale in \citet{papadopoulos2024arrows}, we hypothesize that the ease with which the language model can approximate the factorized distribution of L2R and R2L, may be also tied to which direction exhibits higher branching factors in that direction. We leave this exploration for future study.

\begin{table*}[!htp]\centering
\caption{
Results of the controlled simulation study of 4-digits multiplication. Theoretical Conditional Entropy (Theo. Cond. Ent.) represents the expected conditional entropy under an ideal model. L2R consistently outperforms R2L in Forward X, while R2L is superior in Reverse X. Lower conditional entropy correlates with higher accuracy. 
}\label{tab:sim_results}
\scriptsize
\begin{tabular}{lcccHHccc}
\toprule
&\multicolumn{3}{c}{\textbf{Forward X}} &\multicolumn{2}{H}{Unique X} &\multicolumn{3}{c}{\textbf{Reverse X}} \\\cmidrule{2-4} \cmidrule{7-9}
&L2R &R2L(m,n) &R2L(m) &L2R &R2L &R2L &L2R(m,n) &L2R(n) \\\midrule
Test Accuracy (\%) & \textbf{99.81}$\pm$0.15 & 59.71$\pm$1.99 & 60.93 $\pm$ 0.88&\textbf{96.97}$\pm$0.48 & 53.91$\pm$1.09  & \textbf{100}$\pm$0 & 97.82$\pm$0.35 & 99.85$\pm$0.10\\
Train Accuracy (\%) & \textbf{99.76}$\pm$0.15 & 59.03 $\pm$ 1.66& 61.22$\pm$1.12& 95.84$\pm$0.46& 54.79 $\pm$1.32& \textbf{100}$\pm$0 & 97.90$\pm$0.42 & 99.98$\pm$0.04\\
\midrule
Test Cond. Ent. (nats) & 0.06 & 1.18 & 0.08& 1.04 & 1.50 & 0 & 0.84 & 0.01 \\
Train Cond. Ent. (nats) & 0.06 & 1.17 & 0.08 & 1.05& 1.51 & 0 & 0.83 & 0.01\\
Theo. Cond. Ent. (nats) & 0 & 1.49 & 0 & 0 & 0 & 0 & 1.49 & 0 \\
\midrule
Training loss & \textbf{0.86} & 0.94 & 0.94 & 0.85 & 0.92 & \textbf{0.86} & 0.94 & 0.94\\
\bottomrule
\end{tabular}
\end{table*}

\section{Controlled Simulation Study}
\label{sec:sim}
\begin{wrapfigure}{r}{0.5\textwidth}
    \centering
    \includegraphics[width=\linewidth]{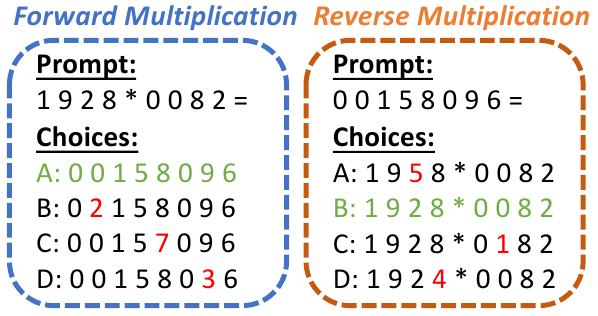}
    \caption{Simulation Study. Forward multiplication simulates a \textbf{many-to-one} mapping scenario, while reverse multiplication simulates a \textbf{one-to-many} mapping.}
    \label{fig:sim}
\end{wrapfigure}

The three hypotheses discussed in Section~\ref{sec:why} are intricately entwined in actual MCQs, making it challenging to disentangle them.
To better investigate the hypotheses explaining the optimal direction for MCQs, we conducted a meticulously controlled simulation study (Figure~\ref{fig:sim}) focused on 4-digit multiplication. Although the arithmetic dataset is different from the real language modeling datasets, this simulation study can be a good controlled experiment to understand the phenomenon we observed in Section~\ref{sec:why}, as we are investigating the underlying principle of the model with different factorizations regardless of the dataset.
The L2R and R2L models were initialized \textbf{from scratch} and \textbf{exclusively} trained on this simulation dataset to eliminate any potential confounding factors. All data instances share the same format and length, removing the \textit{calibration} effect from the analysis and allowing us to concentrate on \textit{computability} and \textit{conditional entropy}.

\paragraph{Experiment Setup}
We conduct two types of simulation experiments: Forward Multiplication (\textbf{Forward X}) and Reverse Multiplication (\textbf{Reverse X}). In Forward X, each training instance was represented as $m \times n = p$, where $m, n \in \{0, \ldots, 10^4\}$ and $p \in \{0, \ldots, 10^8\}$. The formatting included spaces between digits and mathematical operators to ensure a consistent single tokenization for both L2R and R2L models.
In Reverse X, the multiplication was in reverse order, such as $p = m \times n$. For each simulation type, L2R and R2L models were trained with a 2B model size with 1 epoch on all $10^8$ non-repeating equations except 1,000 test examples, totaling almost 3.2B tokens.

The model performance was assessed using the held-out test set with 1,000 examples. These examples were converted into a multiple-choice format consisting of 4 choices (Figure~\ref{fig:sim}). Other than the correct answer, the remaining three hard negative options were created by altering a single digit in the correct answer to a random other digit, at a random position. The presenting order of the four choices are then randomly shuffled. We augmented the test set 10 times and calculated the average accuracy and conditional entropy.

As multiple pairs of $m$ and $n$ can be mapped to the same product $p$, Forward X is a \textbf{many-to-one} mapping. The theoretical conditional entropy for predicting the correct $p$ from $m \times n$ is 0 under an oracle model. However, as there are several paths from the product $p$ to the $m,n$ pairs, the theoretical conditional entropy for predicting the $m \times n$ from $p$ becomes 1.49 nats under an oracle model.
In the Reverse X task, which transitions into a \textbf{one-to-many} scenario, the analysis is inverted.

For Forward X, we explore an alternative R2L evaluation method, denoted as \textbf{R2L(m)}, where the relevance score of the i-th choice $p_i$ is calculated as $s_i=\log p_{R2L}(m \mid p_i, n)$, focusing on the conditional entropy of $m$ rather than $m \times n$ as in the standard \textbf{R2L(m,n)} method. Since R2L(m) is essentially division, it is deterministic with a theoretical conditional entropy of 0. Similarly, we have a variant for L2R in reverse X, called \textbf{L2R(n)}.

\paragraph{Results}
The results are presented in Table~\ref{tab:sim_results}. In Forward X scenarios, L2R models demonstrate higher accuracies than R2L(m,n) models, with correspondingly lower conditional entropy and training loss. This observation aligns with our hypothesis in Section~\ref{sec:why}.
Conversely, in Reverse X scenarios, the R2L model outperforms the L2R(m,n) model. The training and test performance gaps are minimal.

Interestingly, R2L(m) achieves better accuracy than R2L(m,n) in Forward X as conditional entropy decreases. Similarly, L2R(m,n) surpasses L2R(n) in Reverse X. This suggests that \textbf{when maintaining the same thinking direction -- where computability should remain equivalent -- performance improvements can be achieved} by configuring $s_i$ to have lower conditional entropy. 
This hints that the R2L performance on MCQs can potentially be further improved by configuring the input to predict fewer tokens in the question $q$, so that the minimum conditional entropy is obtained.
We leave this for future exploration. 

On the other hand, comparing L2R with R2L(m), where theoretical conditional entropy equals 0, L2R maintains superiority, indicating that \textbf{computability likely remains as a key factor}. 
For the Reverse X task, the accuracy gap between R2L and L2R(n) is smaller than the accuracy gap between L2R(m,n) and L2R(n), suggesting that the conditional entropy may explain more of the performance gap than the computability.

Notably, models achieve higher accuracies on Reverse X compared to their Forward X counterparts, despite similar training loss and conditional entropy values. This disparity could probably be attributed to the closer proximity of choices in Forward X, which inherently increases task difficulty. We provide additional discussion and analysis comparing Forward X and Reverse X in Appendix~\ref{app:simulation}.

\section{Related Work}
\textbf{Reversal Curse} 
\citet{berglund2023reversal} first investigates the "reversal curse" in LLMs, which refers to the phenomenon where models trained on forward text data struggle to perform well on inverse search tasks. \citet{allen2023physics_3_1} further discusses this issue and proposes that augmentation during the pretraining stage can help bridge the knowledge extraction performance gap in reverse entity mapping. In a similar vein, \citet{golovneva2024reverse} suggest training a unified model that combines text data with augmented reversed or partially reversed data can mitigate the reversal curse.
These studies imply that autoregressively-trained language models tend to have a linear and unidirectional thinking process, and certain types of augmentation can facilitate the model in making complex connections between pieces of learned information to enable more intricate cross-referencing. Our research also demonstrates that the autoregressive nature of LLMs may introduce inductive biases rooted from the pretraining corpus. Instead of focusing on the "reversal curse," we suggest that knowledge extraction and reasoning may be more straightforward in the direction with lower conditional entropy.

\textbf{Order of Reasoning} 
Previous work has also explored the reasoning order's impact on reasoning performance. \citet{vinyals2015order} first demonstrates that the sequence in which input and output data are organized significantly impacts the performance of sequence-to-sequence models and propose to search over possible orders during training to manage unstructured output sets.
Recently, \citet{papadopoulos2024arrows} reveals a surprisingly consistent lower log-perplexity when predicting in L2R versus R2L, despite theoretical expectations of symmetry. The authors attribute this asymmetry to factors like sparsity and computational complexity. We also observe this difference yet we have another hypothesis rationale beyond theirs. 
\citet{zhang2024reverse} shows that by reversing the digit order, prioritizing the least significant digit can improve LLMs's performance on arithmetic, which aligns with our findings in section~\ref{sec:sim}.

Previous studies on sequence modeling have also delved into relaxing the conventional ``left-to-right'' autoregressive dependencies, primarily to facilitate parallel generation \citep{gu2018non, ghazvininejad-etal-2019-mask,gu-kong-2021-fully,zhang-etal-2020-pointer} and non-monotonic  generation~\citep{welleck2019non,gu2019insertion}. Text diffusion has recently emerged as a promising approach in terms of planning and controllability \citep{Li-2022-DiffusionLM, zhang2023planner, gong2024scaling}. It has shown to be more effective than language models (LLMs), particularly for tasks that require bidirectional reasoning strategies such as sudoku and countdown games \citep{ye2024beyond}. Alternatively, the Belief State Transformer (BST) \citep{hu2025belief} enhances sequence modeling by using both prefix and suffix inputs to predict subsequent and preceding tokens, effectively capturing a compact belief state for improved goal-conditioned decoding and test-time inference. In contrast, our work primarily investigates the optimal reasoning order for non-generative tasks requiring structured inference.

\textbf{Multiple-Choice Questions (MCQs) for LLM evaluation}
MCQs have been widely used for evaluating LLM's reasoning and knowledge extraction abilities. 
\citet{zheng2023large} demonstrates that LLMs exhibit a selection bias in MCQs, favoring certain option positions, and introduces a debiasing method to mitigate this issue.
\citet{pezeshkpour2023large} examines how LLMs' performance on MCQs is influenced by the order of answer options, finding that reordering can lead to huge performance variations. 
\citet{ghosal2022two} proposes reframing MCQs as a series of binary classifications, demonstrating that this approach significantly improves performance across various models and datasets.
\citet{li2024can} highlights issues like positional biases and discrepancies compared to long-form generated responses, when using MCQs in evaluating LLMs.
\citet{wiegreffe2024answer} discovers that the prediction of specific answer symbols is primarily attributed to a single middle layer's multi-head self-attention mechanism, with subsequent layers increasing the probability of the chosen answer in the model's vocabulary space.
In contrast to the previous work, our work first shows the connection between the preferred reasoning direction and the direction that has lower conditional entropy in MCQ evaluations.

\section{Limitations}

\textbf{Task Scope and Generalizability}
While our findings demonstrate the benefits of R2L factorization in some MCQs, the applicability to other NLP tasks remains unexplored. Our focus on MCQs was intentional to provide a controlled setting for studying factorization effects, as evaluation on generative tasks is not as straightforward. While exploring similar observations in other generative tasks is a valuable avenue for future research, in this paper we limit our discussion to MCQs as a lens to investigate the machinery of LLMs. Additionally, our experiments were limited to models in the 2B-8B parameter range. Due to resource constraints, we were not able to pretrain a SOTA-scale LLM from scratch; future work should explore whether our findings generalize to SOTA models.

\textbf{Theoretical Framework}
We propose three hypotheses (calibration, computability, and conditional entropy) to explain different factorization directions' effectiveness, but our analysis in natural language datasets remains qualitative as it is challenging to measure these factors in real language. We used synthetic arithmetic tasks to better understand factors contributing to the L2R and R2L performance gap. Future work should develop rigorous quantitative measures and establish formal theoretical connections between these factors through more sophisticated controlled experiments.

\textbf{Experimental and Model Design}
Our evaluation methodology, particularly choosing between $p(q|a_i)$ and $p(a_i|q)$ paradigms, may introduce confounding factors that make isolating pure factorization effects challenging. Future work should develop more robust frameworks that better disentangle these effects from task reformulation. While we focus on standalone L2R and R2L models, incorporating both factorizations through a unified model may further improve performance.

\section{Conclusion}
In this work, we investigated what makes the preferred thinking direction for LLMs in multiple-choice questions. Through extensive experimentation with models of varying sizes and training datasets, we discovered the surprising finding that R2L factorization can outperform traditional L2R approaches in specific MCQ tasks (4 out of 11 evaluated benchmarks). Our analysis revealed that the effectiveness of each factorization direction may be intrinsically linked to several factors including calibration, computability, and conditional entropy of the downstream task distribution, with lower conditional entropy yielding better performance. We disentangled and validated these factors through controlled simulation studies using arithmetic tasks.

The core contribution of this work lies in bringing this phenomenon to the community's attention and initiating analysis into the underlying factors that might explain why a particular thinking direction is preferred in specific task domains. These findings may suggest the potential for future language model development by revealing the knowledge extraction and reasoning machinery of LLMs and suggesting that alternative factorizations deserve serious consideration in model design. Future work could explore additional factorization strategies beyond L2R and R2L, investigate applications to other types of language tasks, and develop more sophisticated methods for combining different factorizations.

\input{sections/ack}

\bibliography{arxiv}
\bibliographystyle{iclr2024_conference}

\newpage
\input{sections/appendix}

\end{document}

%% file: math_commands.tex
\usepackage{amsmath,amsfonts,bm}

\def\eqref#1{equation~\ref{#1}}

\def\1{\bm{1}}

\DeclareMathAlphabet{\mathsfit}{\encodingdefault}{\sfdefault}{m}{sl}
\SetMathAlphabet{\mathsfit}{bold}{\encodingdefault}{\sfdefault}{bx}{n}

%% file: sections/ack.tex
\section*{Acknowledgment}
We thank Iman Mirzadeh, Mehrdad Farajtabar, Stephen Pulman, Tatiana Likhomanenko, Dan Busbridge, Jason Ramapuram, Huangjie Zheng, Shansan Gong, Shuangfei Zhai, Jiarui Lu and Jiasen Lu for their invaluable insights and discussions that have greatly benefited our project.

%% file: sections/appendix.tex
\appendix
\onecolumn
\begin{center}
    {\Large \bf Appendix}
\end{center}
\section{Model Architecture and Training Hyperparameters}
\label{app:arch}
We provide the model architecture of our 2B and 8B models in Table~\ref{tab:model_architecture}. We use no attention dropout and the Normalization $\epsilon$ is set to $1 \times 10^{-5}$.  
We employ cosine scheduling for the learning rate, with the number of warmup steps set at 2\% of the total iterations. The models are trained on 64 H100 Nvidia GPUs for 60 hours and 208 hours for 2B and 8B models, respectively.

We used the Llama3 tokenizer \citep{dubey2024llama3}. Despite being trained from a L2R model, \citet{papadopoulos2024arrows} demonstrated through an ablation study that the direction of tokenizer being trained has minimal effect on the R2L training. Consequently, we adhere to the same reasoning and opt not to train on a R2L tokenizer, as obtaining the necessary training data for Llama3's tokenizer is practically infeasible for fair comparison.
Our code implementation is based on Torchtune \citep{pytorch2024torchtune}, which is released under the BSD 3-Clause License, a permissive open-source license that allows for broad usage and distribution with minimal restrictions. 

\begin{table}[!htp]\centering
\rowcolors{2}{gray!15}{white}
    \caption{Model architecture specifications for 2B and 8B models.}\label{tab:model_architecture}
    \begin{tabular}{lrrrr}\toprule
    Parameter & 1.5B Model & 2B Model & 4B Model & 8B Model \\\midrule
    Vocabulary Size & 128,256 & 128,256 & 128,256 & 128,256 \\
    Number of Layers & 18 & 24 &24 & 32 \\
    Number of Heads & 16 & 16 & 16 & 32 \\
    Number of KV Heads & 16 & 8 & 8 & 8 \\
    Embedding Dimension & 2048 & 2,048 & 3072 & 4,096 \\
    Max Sequence Length & 2,048 & 2048 & 2,048 & 8,192 \\
    Intermediate Dimension & 5,632 & 7,168 & 11,264 &14,336 \\
    RoPE Base & 500,000 & 500,000 & 500,000 & 500,000 \\
    \bottomrule
    \end{tabular}
\end{table}

\section{Results with Different Model Sizes}
\label{app:full_results}
We conducted additional experiments to explore performance trends across a broader range of model scales, training 1B and 4B parameter models (R2L and L2R) to complement our existing 2B and 8B models. 
The full results, presented in Table~\ref{tab:full_results}, consistently reinforce our main claim.

\begin{table}[!htp]\centering
    \rowcolors{2}{gray!15}{white}
    \caption{Results with different model sizes.}\label{tab:full_results}
    \small
    \begin{tabular}{lrrrrrrrrr}\toprule
    &\multicolumn{2}{c}{EDU-1.5B} &\multicolumn{2}{c}{EDU-2B} &\multicolumn{2}{c}{EDU-4B} &\multicolumn{2}{c}{EDU-8B} \\\cmidrule{2-9}
    &L2R &R2L &L2R &R2L &L2R &R2L &L2R &R2L \\\midrule
    \textbf{LogiQA} &28.10 &\textbf{32.65} &27.96 &\textbf{31.49} &26.58 &\textbf{32.72} &29.95 &\textbf{31.03} \\
    \textbf{OpenbookqQA} &38.00 &\textbf{41.20} &42.40 &\textbf{44.40} &43.00 &\textbf{44.60} &45.00 &\textbf{48.40} \\
    \textbf{TruthfulQA} &23.01 &\textbf{28.86} &24.36 &\textbf{28.76} &22.40 &\textbf{29.00} &24.97 &\textbf{31.70} \\
    \textbf{CommensenseQA} &41.44 &\textbf{44.22} &42.92 &\textbf{45.13} &\textbf{50.70} &43.40 &39.15 &\textbf{44.96} \\
    Social\_IQA &41.35 &\textbf{41.65} &\textbf{42.78} &42.22 &\textbf{42.99} &40.00 &\textbf{44.58} &43.50 \\
    AI2\_arc &\textbf{58.63} &51.8 &\textbf{60.65} &52.31 &\textbf{63.14} &53.95 &\textbf{68.29} &56.22 \\
    HellaSwag &\textbf{56.46} &43.38 &\textbf{60.57} &44.34 &\textbf{65.93} &46.38 &\textbf{71.60} &49.22 \\
    MathQA &\textbf{25.52} &24.05 &\textbf{26.80} &24.86 &\textbf{27.33} &24.08 &\textbf{28.77} &25.33 \\
    MMLU &33.30 &\textbf{33.40} &\textbf{34.57} &34.35 &\textbf{36.70} &35.16 &\textbf{38.90} &37.11 \\
    PIQA &\textbf{73.23} &59.19 &\textbf{74.48} &57.13 &\textbf{76.00} &58.76 &\textbf{77.80} &59.14 \\
    Winogrande &\textbf{58.56} &51.77 &\textbf{60.93} &54.85 &\textbf{61.80} &54.50 &\textbf{65.75} &54.70 \\
    \bottomrule
    \end{tabular}
    \end{table}

Specifically, we observe a clear scaling trend: for the tasks where reverse thinking excels (LogiQA, OpenbookQA, TruthfulQA, and Commonsense QA), the performance advantage of R2L over L2R grows as the model size increases. 
Although we did not scale beyond 8B parameters due to engineering constraints, these extended results demonstrate the robustness of our findings. 
This trend strongly suggests that the impact of alternative factorizations is unlikely to diminish with even larger models; on the contrary, their benefits may become more significant.

\section{Comparing Three Paradigms for Reverse Thinking}
\label{app:rev_comparison}
We present a comparison of the three paradigms discussed in Section~\ref{sec:rev} across all tasks, as shown in Table~\ref{tab:3variants}. Paradigm 3 showcases the highest overall performance.

\begin{table}[!htp]\centering
\rowcolors{2}{gray!15}{white}
    \caption{Downstream evaluation of three variants to compute $s_i$. }\label{tab:3variants}
    \small
    \begin{tabular}{lrrr}\toprule
    &Paradigm 1 &Paradigm 2 &Paradigm 3 \\\midrule
    LogiQA &31.49\% &27.34\% &\textbf{31.49\%} \\
    OpenbookqQA &37.00\% &24.60\% &\textbf{44.40\%} \\
    TruthfulQA &24.36\% &17.38\% &\textbf{28.76\%} \\
    CommensenseQA &41.85\% &27.19\% &\textbf{45.13\%} \\
    Social\_IQA &\textbf{43.09\%} &37.36\% &42.22\% \\
    AI2\_arc &32.19\% &45.32\% &\textbf{52.31\%} \\
    HellaSwag &33.73\% &43.38\% &\textbf{44.34\%} \\
    MathQA &23.18\% &21.07\% &\textbf{24.86\%} \\
    MMLU & 33.10\% &29.31\% &\textbf{34.35\%} \\
    PIQA &\textbf{70.76\%} &68.70\% &57.13\% \\
    Winogrande & 54.46\% &53.67\% &\textbf{54.85\%} \\
    AVG &38.66\% &35.94\% &\textbf{41.80\%} \\
    \bottomrule
    \end{tabular}
    \end{table}

\section{Experimental Details on Computing the Conditional Entropy in MCQs}
\label{app:ce}
We measure conditional entropy across various MCQs tasks by using ancestral sampling to generate sequences conditioned on each task's input. Generation length can be a confounding issue as longer generation will typically receive larger 
entropy. 
To ensure consistent and fair comparison, we constrain the generation to exactly 10 tokens in each direction (L2R and R2L). We make the generation non-stopping by removing the end-of-sentence token. The conditional entropy is first calculated for individual examples within each task, then averaged across all examples to obtain the task-level measure.

\section{Analysis of How Reverse Thinking Reduces Surface Form Competition in MCQ}
\label{app:surface}
Consider a multiple-choice question:

\begin{center}
    Q: \textbf{Who barks?} \\
    A: \textit{dog} \quad B: \textit{cat}
\end{center}

We assume a vocabulary of only three words: \textit{dog}, \textit{cat}, and \textit{puppy}. However, the answer choices are limited to \textit{dog} and \textit{cat}.

\paragraph{Forward Thinking: Surface Form Competition}
In the forward model, suppose the model predicts the probabilities of each token in the vocabulary as in below:
\[
P_{L2R}(\text{dog} \mid q) = 0.3, \quad P_{L2R}(\text{puppy} \mid q) = 0.3, \quad P_{L2R}(\text{cat} \mid q) = 0.4.
\]
As \citet{holtzman-etal-2021-surface} shown, here, ``dog'' and ``puppy'', although referring to the same entity, compete as surface forms, potentially impacting the selection of ``dog'' as an answer.

\paragraph{Reverse Thinking: Alleviating the Surface Form Competition}
In the reverse thinking, the model should still suffer the surface competition in the prior, suppose we have:
\[
P_{R2L}(\text{dog}) = P_{R2L}(\text{puppy}) = 0.2, \quad P_{R2L}(\text{cat}) = 0.6.
\]

The reverse conditional probability would would resemble this:
\[
P_{R2L}(q \mid \text{dog}) = 0.9, \quad P_{R2L}(q \mid \text{puppy}) = 0.9, \quad P_{R2L}(q \mid \text{cat}) = 0.4.
\]

In the reverse thinking R2L paradigm 3, we enforce a uniform prior, which achieves the best performance. 
We hypothesize that this is due to the alleviation of the surface form competition issue. In this example, by enforcing a uniform prior, surface form competition in the prior distribution disappears, ensuring a fair comparison between ``dog'' and ``cat''.
Since we only compare $P_{R2L}(q \mid \text{dog})$ and $P_{R2L}(q \mid \text{cat})$, the probability $P_{R2L}(q \mid \text{puppy})$ becomes irrelevant.

\section{Additional Results for Simulation Study}
\label{app:simulation}
We think Forward X is generally more challenging because distinguishing incorrect choices from the correct one can be more challenging than in the Reverse X. Unlike Reverse X where all wrong choices are seen in the dataset, approximately 75\% of incorrect choices in Forward X are absent as they cannot be factorized into $m \times n$. 

We provide some additional results by performing free ancestral sampling generation (without topK or topP, with temperature set to 1) for the R2L(m) in Forward X. The accuracy of exact match for the generation is reported in below. For reference, the L2R free generation accuracy is 90.4\%.

\begin{enumerate}
    \item Regarding Forward X, R2L(m), free generation of $m$ using $\times n = p_{\text{correct}}$, the exact match accuracy for generative output is \textbf{82.1\%}.
    \item For Forward X, R2L(m), free generation of $m$  using $\times n = p_{\text{incorrect}}$, the generative exact match accuracy remains high at \textbf{30.2\%}, while the accuracy for random guessing is nearly 0. 
\end{enumerate}

These results suggest that the R2L model may map unseen $p_{\text{incorrect}}$ to its neighborhood $p_{\text{correct}}$ and execute the same computation as if using $p_{\text{correct}}$. Consequently, the model faces a significant challenge in distinguishing alternative incorrect answers from the correct ones, making alternative incorrect answers hard-negatives. 

\section{Statistical Testing Results}
\label{app:ttest}
We presents detailed statistical comparisons between L2R and R2L reasoning approaches across four tasks where R2L outperforms L2R: LogiQA, OpenBookQA, TruthfulQA, and CommonsenseQA. Metrics include mean accuracy, standard deviation, and the results of a paired t-test comparing the two approaches. All results were computed using bootstrap sampling with five replicates, each resampled at 80\% with replacement, providing a robust estimation of performance variance.
Notably, R2L consistently outperformed L2R, with statistically significant improvements ($p$-value < 0.05) observed across all tasks. The largest relative performance gain appeared in LogiQA, where R2L achieved a mean accuracy increase from 27.04\% (L2R) to 32.56\% (R2L) with an extremely significant $p$-value. OpenBookQA also exhibited a clear improvement, rising from 42.80\% to 44.40\%. TruthfulQA and CommonsenseQA saw meaningful gains as well, with $p$-values of 0.0072 and 0.0471, respectively, indicating reliable improvements beyond random fluctuation.
Overall, these results underscore the effectiveness of reverse thinking (R2L) for improving reasoning tasks.

\begin{table}[!htp]\centering
\rowcolors{2}{gray!15}{white}
    \caption{Paired t-test of bootstrap samples (5 replicates, 80\% resampled with replacement).}\label{tab:ttest}
    \begin{tabular}{lccccc}\toprule
    Task & L2R Mean & L2R Std & R2L Mean & R2L Std & $p$-value \\\midrule
    LogiQA & 27.04 & 0.00 & 32.56 & 0.0045 & 0.0000  \\
    OpenBookQA & 42.80 & 0.00 & 44.40 & 0.0000 & 0.0000 \\
    TruthfulQA & 25.08 & 1.16 & 28.93 & 0.80 & 0.0072 \\
    CommonsenseQA & 42.17 & 1.34 & 44.68 & 1.70 & 0.0471  \\
    \bottomrule
    \end{tabular}
\end{table}